\documentclass[final]{cvpr}

\usepackage{times}
\usepackage{epsfig}
\usepackage{graphicx}
\usepackage{amsmath}
\usepackage{amssymb}
\usepackage{caption}

\usepackage[pagebackref=true,breaklinks=true,colorlinks,bookmarks=false]{hyperref}

\def\cvprPaperID{****} % *** Enter the CVPR Paper ID here
\def\confYear{CVPR 2025}

\begin{document}

%%%%%%%%% TITLE
\title{SketchColour: Channel Concat Guided  DiT-based \\Sketch-to-Colour Pipeline for 2D Animation}

\author{Bryan Constantine Sadihin\quad Michael Hua Wang\quad Shei Pern Chua\quad Hang Su\\
Department of Computer Science, Tsinghua University\\
{\tt\small \{wangwd24,\,wanghua24,\,cxp24\}@mails.tsinghua.edu.cn,\; suhangss@mail.tsinghua.edu.cn}
}

\twocolumn[{
\maketitle
  \vspace{-1em}            % tighten up if you like
  \centering
  \includegraphics[width=0.7\linewidth]{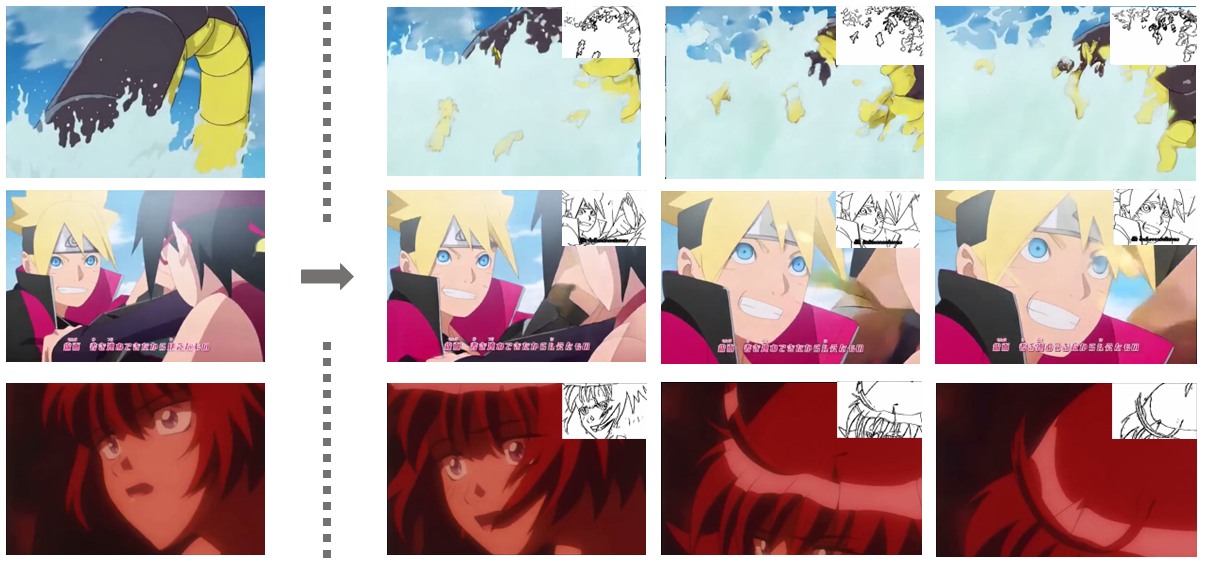}
  \captionof{figure}{%
    \textbf{SketchColour} receives the colored first frame and the entire scene in sketch format, then colors each frame based on the reference.}
  \label{fig:opening_figure}
  \vspace{1em}            % add a bit of space before the two-column text
}]
%%%%%%%%% ABSTRACT
\begin{abstract}
   The production of high-quality 2D animation is highly labor-intensive process, as animators are currently required to draw and color a large number of frames by hand. We present SketchColour, the first sketch-to-colour pipeline for 2D animation built on a diffusion transformer (DiT) backbone. By replacing the conventional U-Net denoiser with a DiT-style architecture and injecting sketch information via lightweight channel-concatenation adapters accompanied with LoRA finetuning, our method natively integrates conditioning without the parameter and memory bloat of a duplicated ControlNet, greatly reducing parameter count and GPU memory usage. Evaluated on the SAKUGA dataset, SketchColour outperforms previous state-of-the-art video colourization methods across all metrics, despite using only half the training data of competing models. Our approach produces temporally coherent animations with minimal artifacts such as colour bleeding or object deformation.
   
  Our code is available at: \url{https://bconstantine.github.io/SketchColour/}.
\end{abstract}

%%%%%%%%% BODY TEXT
\section{Introduction}

\begin{figure}[t]
\begin{center}
\includegraphics[width=1.0\linewidth]{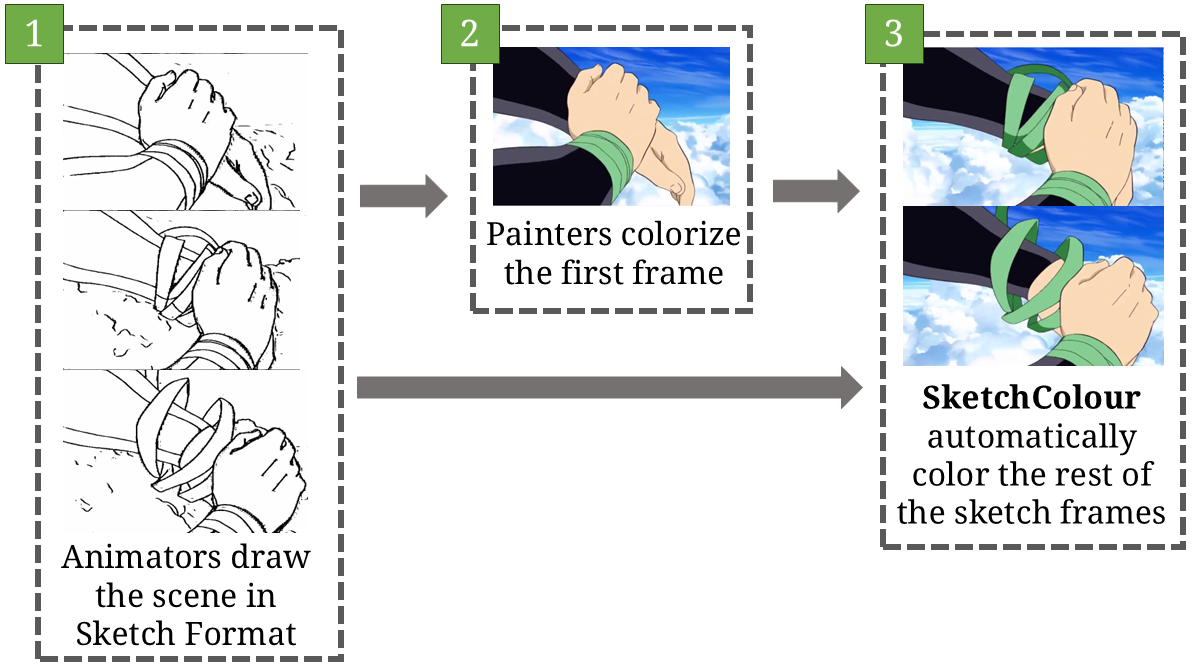}
\end{center}
   \caption{Due to the frame-by-frame workflow of 2D animation, painters must meticulously add color frame-by-frame. SketchColour helps animators by automating the colorization of subsequent frames following the reference color given by the first frame. }
\label{fig:explaining_usage}
\label{fig:onecol}
\end{figure}

The production of high-quality 2D animation is a labor-intensive task. Artists must meticulously draw each frame through successive stages of sketching the main object and colorizing the sketch. \cite{Anita2024} (see Figure \ref{fig:explaining_usage}). While this process permits precise artistic control, it also imposes significant time and labor and costs on animation studios, slowing down content pipelines and limiting creative iteration \cite{meng2024anidoc}. Generating in-between frames with an initial product prototype, such as a sketched version of the video, using an automated system allows studios to streamline their workflows, thus accelerating delivery of animated content to meet growing audience demand without sacrificing any fine-grained animation controllability.

Recent innovation in diffusion-based image-to-video (I2V) models have demonstrated impressive capabilities for generating short video clips from static images guided by text instructions. Controllable I2V is a subdomain of I2V where finer input control is added to direct the generated video details with the usage of control modalities such as trajectory points \cite{wang2024framer}, reference videos \cite{ling2024motionclone}, or bounding boxes and masks \cite{li2025magicmotion}.  However, the aforementioned control modalities sufficiently support the animator’s need for explicit, fine-grained control over animation details. In contrast, using the clean line-art sketch itself as the control signal lets artists keep drawing exactly as they do today: by colouring only the first keyframe, they may feed their remaining sketches to the model, and receive a fully coloured sequence. This promises a faster creation process while not sacrificing control over fine-grained details during the creation process.

Early works attempt sketch colourisation as a frame-level task, applying GANs or U-Net diffusion models to each frame independently \cite{li2022eliminating, yan2025colorize, cao2023animediffusion}. However, these approaches have issues with color consistency across the sequence (\eg frame flickering) and also propagate colouring error. More recent systems have moved to video diffusion, but still inherit limitations \cite{meng2024anidoc, xing2024tooncrafter, huang2024lvcd}. Past models usually use a U-Net based model \cite{blattmann2023svd, he2022lvdm} with an extra sketch-guided ControlNet \cite{zhang2023controlnet}, which requires replication of the model architecture in part or in whole. This bloats the trainable parameter count and risks convergence instability due to juggling two identical but separately-updated networks. Additionally, even when given the first frame as reference, ControlNet is also known to have issues with "latent-gap artefacts" \cite{yan2025colorize} in cases where the coloured image and subsequent sketches occupy different latent manifolds, \ie when the dense RGB context and sparse line-art's pixel-wise alignment diverge. This is encountered in animation colorization when later frames' structure strays away from that of the colored first frame, which is reference. Due to this imbalanced injection, prior U-Net + ControlNet pipelines report colour bleed when processing sequences with vigorous motion. Furthermore, as the U-Net backbone adopted by prior work first down-samples feature maps through several resolution stages before any global attention is applied, fine-grained sketch details, in this case colorization detail, is often compromised.

These constraints motivate a backbone that can: fine-tune conditional guidance without parameter bloat, natively integrate condition information, and enhance fine-grained colorization ability. For that reason, we propose SketchColour, the first diffusion transformer (DiT) framework tailored to sketch-conditioned animation colourisation. The model replaces the U-Net denoiser with a diffusion transformer-style \cite{Peebles2022DiT} backbone and injects the sketch signal via lightweight channel-concatenation adapters, eliminating the need for a separate ControlNet.

Our contributions are as follows:
\begin{itemize}
    \item SketchColour presents the first sketch-to-colour pipeline for 2D animation built on a DiT backbone. Due to its ability to understand global context, our method outperforms traditional U-Net diffusion approaches in both fidelity and consistency.
    \item Utilizing Channel Concat Control combined with fine-tuning a small LoRA of only 10 million parameters (compared to the billions of parameters used by ControlNet), our parameter-efficient finetuning technique minimizing both the required number of training steps and the size of the training dataset. 
    \item Our work defeats all previous state-of-the-art model in the sketch colorization task, SketchColour enables finer colored guidance and minimizes latent gap artifacts such as color bleeding. 
\end{itemize}  

\begin{figure*}
\begin{center}
\includegraphics[width=0.75\linewidth]{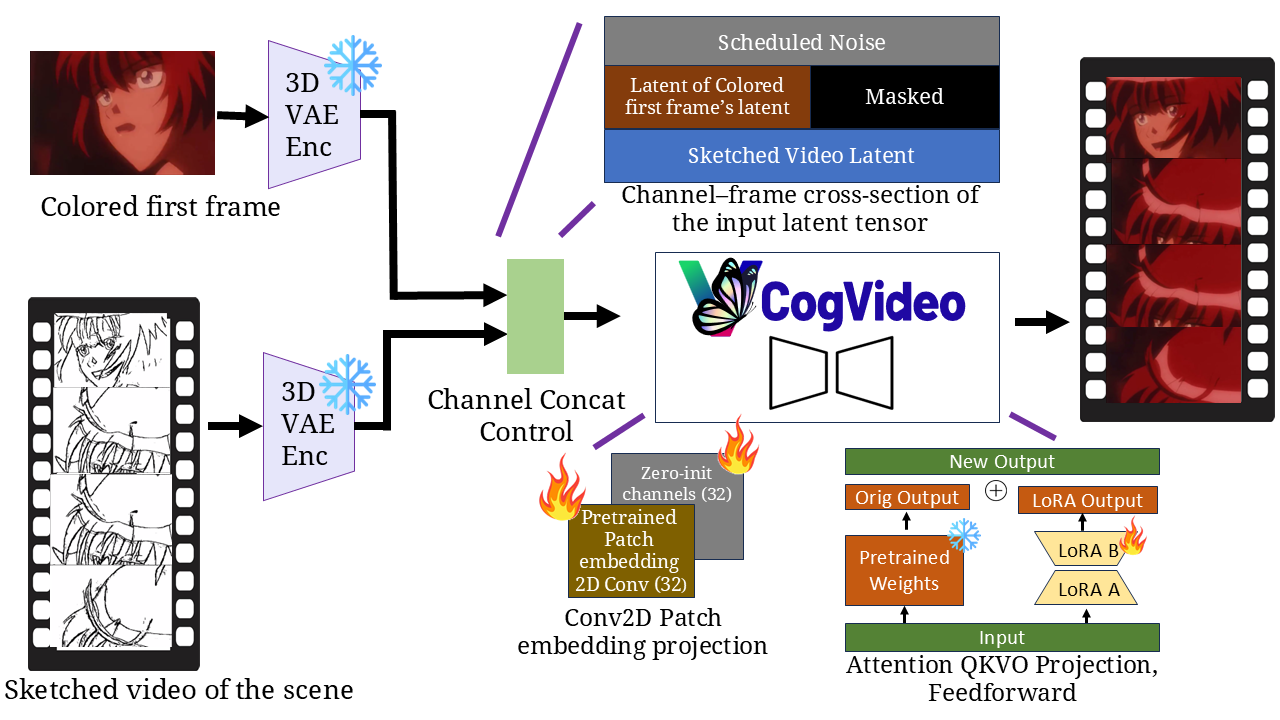}
\end{center}
   \caption{The SketchColour model pipeline. Our model uses a frozen VAE to encode both the colored first frame and the sketches of following frames. We concatenate the latents of these inputs channel-wise and feed to CogVideoX. We fine-tune on the patch embedding projection by expanding the new channel with zero-initialized weights, following ControlNet's approach, and fine-tune the projection \& feedforward layer of attention with LoRA weights. }
\label{fig:model_pipeline}
\end{figure*}

%------------------------------------------------------------------------
\section{Related Work}
\subsection{Diffusion Model Architecture}
Recent advances in video generation models are powered by modifications to the transformer-based architecture for those models. Older models like LVDM \cite{he2022lvdm} and SVD \cite{blattmann2023svd} use a convolutional U-Net architecture as its main diffusion backbone. However, notable discrepancies in quality are visible compared to outputs generated by models such as OpenAI's closed-source video generation model SORA \cite{liu2024sora} utilizing a diffusion transformer (DiT) \cite{Peebles2022DiT} architecture. These DiT models tokenize the video latents, previously compressed by 3D VAE, into spatiotemporal patches and operates on these patches with a transformer based architecture. Motivated by SORA's architecture performance, modern open-source models are typically built upon DiT \cite{yang2024cogvideox, wan2025, kong2024hunyuanvideo}. Despite this advancement, existing models rely solely on text or image guidance for video generation, offering limited control over fine-grained event details, which is essential for animators and other creative industry professionals.
\subsection{Image-level Sketch Colorization}
Traditional methods of image-level sketch colorization rely on GAN-based architectures \cite{kim2019tag2pix} to handle colorization. However, this approach tends to create flat, blotchy regions of color, as the U-Net backbone only sees local context. Further methods use image diffusion models to increase fidelity, \eg ColorizeDiffusion \cite{yan2025colorize} which fine-tunes Stable Diffusion for line art. However, these models colorize each frame independently, causing observable flickering due to slight differences in hue between frames. Recent image colorization works add multi-reference image colorization with diffusion \cite{zhang2025magiccolor, zhuang2025cobra}, which report that ControlNet has issues when the references provided are not sufficiently similar in structure to the target sketch (the latent gap problem). These issues are further demonstrated by the colour bleed effect which occurs during video colorization. This highlights the need for lighter or better-aligned conditioning.
\subsection{Reference-based Video Colorization}
Applying video diffusion models to generate animation output minimizes temporal flickering. ToonCrafter \cite{xing2024tooncrafter} fine-tunes spatial layers of DynamiCrafter \cite{xing2023dynamicrafter}, a U-Net based model, and adds an additional sketch encoder trained on an animation dataset, allowing ToonCrafter to support sketch colorization as an additional use case. However, not only does ToonCrafter require two colored images to support colorization (both sequence start and sequence end), thus requiring the animator to generate additional inputs, but it also struggles to model object motion beyond static shots, commonly displaying problems such as deformed objects. Concurrently, LVCD \cite{huang2024lvcd} also fine-tuned Stable Video Diffusion \cite{blattmann2023svd}, which also uses a U-Net backbone with a sketch-based ControlNet \cite{zhang2023controlnet} for the sketch colorization task. However, LVCD struggles to accurately color animation, producing "dull" or "washed-out" colors as well as suffering from color bleed due to the aforementioned issues that ControlNet has when the discrepancies between the references and the supplied subjects are too large. AniDoc \cite{meng2024anidoc}, a recent work, has drastically improved colorization results by using Stable Video Diffusion and ControlNet fine-tuning as its base model using motion hints during training. However, although reduced, AniDoc still experiences the same latent-gap discrepancy issue in ControlNet, which causes color bleeding. 

Furthermore, all the aforementioned models use ControlNet, which both bloats the trainable parameter count to billions of parameters and risks convergence instability due to needing to juggle two identical but separately-updated networks.

%-------------------------------------------------------------------------

\section{Method}
Our work focuses on colorization of line art videos with the first frame as a reference (see \autoref{fig:model_pipeline}). Our model receives input in the form of the colored first frame $I_{\text{start}} \in \mathbb{R}^{1xHxWx3}$, defining the color and the style of the image, and a sequence of sketches describing the desired video $S \in \mathbb{R}^{TxHxWx3}$, where $T$ is the length of the video in frames. Our model's objective is to output a video $V \in \mathbb{R}^{TxHxWx3}$ such that:
\begin{enumerate}
    \item The result $V$ is the colorized version of sketch $S$.
    \item The colorization applied to $V$ is consistent with the reference $I_{\text{start}}$, with the resultant colorization being temporally coherent in both style and colorization. 
\end{enumerate}
\subsection{Pipeline Design}
\paragraph{Base I2V model} We utilize CogVideoX-5B-I2V as our base image-to-video model.  Our objective is to utilize modern DiT models which permit a larger attention scope than previous U-Net-based models. Due to limited computation resources (see: \autoref{section:implementation details}), we opted to use pretrained I2V models to take advantage of pre-trained image knowledge. As of the current time, CogVideoX-I2V-5B is the smallest available I2V model in the CogVideoX series of I2V DiT models.

\paragraph{3D VAE Encoder} DiT models are paired with a 3D VAE encoder that projects the spatio-temporal information of the input from Image/Video space into the latent representation. In the case of CogVideoX, the latents of the starting frame and the noise-scheduled ground truth video (or gaussian noise latent during inference) are compressed to $Z_{I_{\text{start}}}, Z_{V_{\text{GT}}} \in \mathbb{R}^{\frac{T}{4}\times\frac{H}{4}\times\frac{W}{4}\times16}$ , where $Z_{I_{\text{start}}}$ is zero-padded to match the length of $Z_{GT}$. These two latents are then concatenated channel-wise before being fed into a diffusion transformer block for deionization and generation of the output latent. The output latent is decoded with the same 3D VAE Encoder to output the generated video $V$.  
\paragraph{Fine-tuning: Channel Concat Control and LoRA} Previous approaches opt to use ControlNet when adding extra control modalities. However, these approaches are both expensive to train due to the high parameter count required to replicate the model structure and prone to learning instability due to imbalanced conditional injection, which appears in cases where the later part of the animation is less similar to the reference than the beginning part. This imbalanced conditional injection causes a recurring problem known as color bleed, where the model incorrectly applies color hints based purely on spatial positioning in the first colored reference frame to objects in the later frames of the animation. 

To solve this problem, we use a straightforward approach of concatenating the sketch modality $Z_{\text{sketch}} \in \mathbb{R}^{\frac{T}{4}\times\frac{H}{4}\times\frac{W}{4}\times16}$ channel-wise with $Z_{I_{\text{start}}}$ and $Z_{V_{\text{GT}}}$. Our method enforces that the latent mapping matches between frames, ensuring that color in the later sketches remains masked and must be inferred by the model. Furthermore, $Z_{\text{sketch}}$ is encoded from the original frozen 3D VAE Encoder, without needing to fine-tune. We show (see \autoref{section:frozen vae encoding}) that the representation of the sketch latent is still intact, even though the sketch is in a different modality from the RGB-space reference image. Then, we fine-tune our model and LoRA on the patch projection layer and attention layers, respectively. For the patch projection, we utilize the pretrained projection weights of $Z_{I_{\text{start}}}$ and $Z_{V_{\text{GT}}}$, and initialize the projection weights of $Z_{\text{sketch}}$ with zero-initialized weights, similarly to ControlNet, while maintaining the magnitude of the output into the transformer component. For the transformer section, we add a LoRA for the Attention QKVO Projection and feed-forward layer.
\subsection{Sketch Generation}

 We use Anime2Sketch \cite{Anime2Sketch} to do frame-level conversion of colorized frames to a sketch equivalent, additionally binarizing the sketch results to avoid color information leakage through color intensity \cite{meng2024anidoc}. We used this model to allow for fair comparison with prior work and, which use sketches with similar characteristics.
%------------------------------------------------------------------------

\begin{table*}[t]
\begin{center}
\begin{tabular}{|l|c|c|c|c|c|}
\hline
\multicolumn{6}{|c|}{\textbf{14 Frames}} \\
\hline
\textbf{Method}               & \textbf{MSCE ↓} & \textbf{PSNR ↑} & \textbf{SSIM ↑} & \textbf{LPIPS ↓} & \textbf{FVD ↓} \\
\hline\hline
AniDoc                         & 2612.61$(\pm 4823.98)$         & 18.04$(\pm 4.99)$           & 0.73$(\pm 0.12)$            & 0.30$(\pm 0.13)$             & 898.19$(\pm 704.30)$         \\
LVCD                           & 7937.33$(\pm 4727.72)$         & 10.00$(\pm 2.75)$           & 0.57$(\pm 0.16)$            & 0.45$(\pm 0.12)$             & 2738.45$(\pm 1555.28)$        \\
\textbf{SketchColour (Ours) }          & \textbf{2214.18 $(\pm 3867.13)$}        & \textbf{20.23 $(\pm 5.82)$}           & \textbf{0.79 $(\pm 0.12)$}           & \textbf{0.24 $(\pm 0.14)$}            & \textbf{829.27 $(\pm 723.77)$}        \\
\hline
\multicolumn{6}{|c|}{\textbf{16 Frames}} \\
\hline\hline
ToonCrafter                    & 4619.98 $(\pm 4086.97)$        & 13.06$(\pm 3.52)$           & 0.56 $(\pm 0.13)$           & 0.47 $(\pm 0.13)$            & 1464.59$(\pm 1030.63)$        \\
\textbf{SketchColour (Ours) }          & \textbf{2403.40 $(\pm 4075.10)$}     & \textbf{19.75 $(\pm 5.83)$}          & \textbf{0.78 $(\pm 0.12)$ }          & \textbf{0.25 $(\pm 0.14)$}            & \textbf{860.78  $(\pm 750.60)$ }     \\
\hline
\multicolumn{6}{|c|}{\textbf{17 Frames}} \\
\hline\hline
\textbf{SketchColour (Ours)}           & \textbf{2512.78 $(\pm 4190.19)$}               & \textbf{19.51 $(\pm 5.83)$}              & \textbf{0.78 $(\pm 0.12)$ }              & \textbf{0.25 $(\pm 0.14)$     }          & \textbf{918.70 $(\pm 771.13)$}             \\
\hline
\end{tabular}
\end{center}
\caption{Quantitative comparison of $mean (\pm std)$ video colorization methods at different frame lengths. We display results at identical frame count as the baselines for fair comparison.}
\label{tab:quant_results}
\end{table*}

\section{Experiment}
\subsection{Implementation Details}
\label{section:implementation details}
Considering both our available computation resources and previous works, whose video length is limited to 14 or 16 frames, we train our CogVideoX to generate clips with a fixed length of 17 frames, the minimum length of videos generated by CogVideoX. We use the SAKUGA dataset \cite{sakuga42m2024}, which is composed of animation video clips split into individual scene with text descriptions generated by BLIP-2 \cite{li2023blip}. We filter out elements of the dataset that were already in sketch format, leaving roughly 150K training videos and 60K test videos. From these remaining videos, we sampled 80K videos from the training set and 1K videos from the test set, choosing clips with 17 frames or more and prioritizing based on the shortest such clips. When selecting clips with more than 17 frames, a single continuous 17 frame sequence was randomly sampled from the full clip.

All of our projects were implemented on 2 NVIDIA A40 GPUs. As CogVideoX requires that videos be 720 x 480, we used a fill-and-crop strategy to enforce that our data was of the appropriate resolution. Due to the our limited computation resources, we used a Lora of rank 192 and sample our training dataset down to 80K samples, performing DDP training for 40K learning steps with a batch size of 2 and with AdamW optimizer set to a learning rate of 1e-4. Training took roughly 4 days to complete.

\subsection{Frozen VAE Encoder Information}
\label{section:frozen vae encoding}
\begin{figure}[t]
\begin{center}
\includegraphics[width=1.0\linewidth]{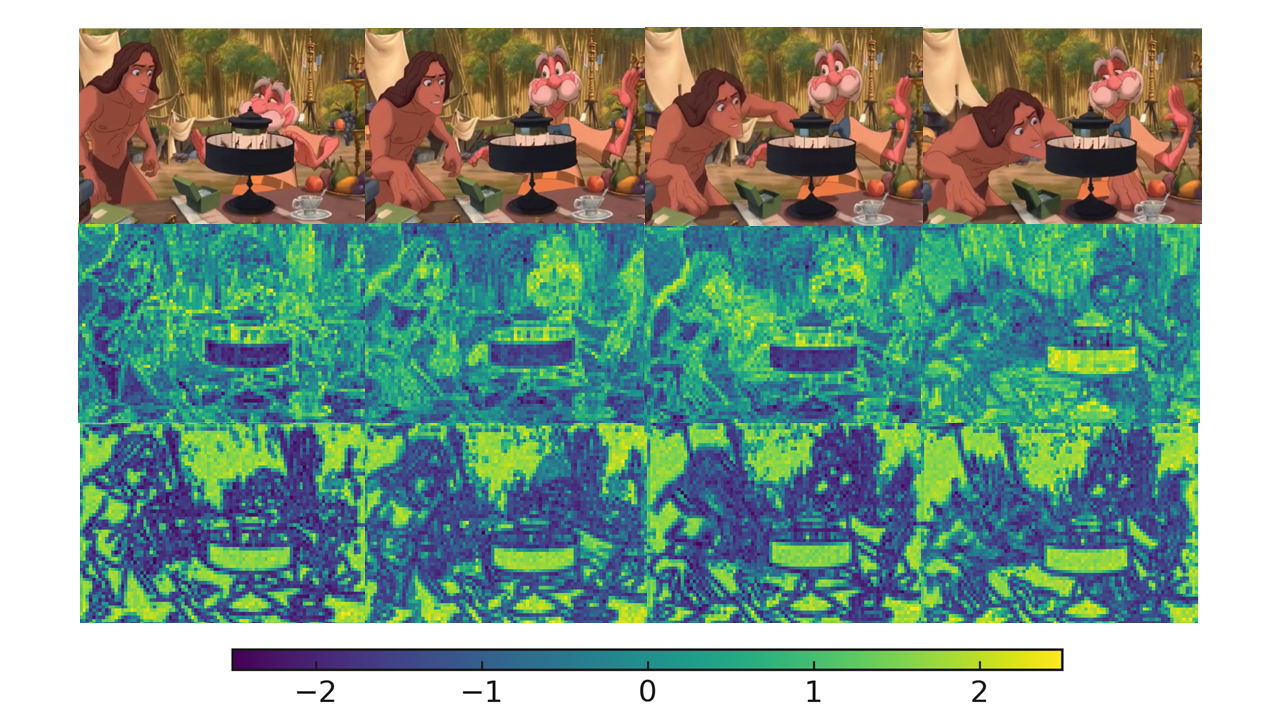}
\end{center}
   \caption{A representation of a video scene on the top row and the corresponding 3D VAE encoded latents for colorized video and sketched video on the middle row and bottom row, repsectively. We apply PCA on the 3D VAE latents on the channel dimension. As can be seen, while the 3D VAE encoded latent carries spatiotemporal information, its spatial representation is still visible. Moreover, the sketched latent representations closely resemble the sketched representation of the colorized video latents, showing that frozen 3D VAE is sufficiently robust for sketch encoding. }
\label{fig:frozen_vae_result}
\end{figure}
We used a frozen 3D VAE encoder to encode the colorized starting frame, ground truth video latent, and corresponding sketch sequence. The behaviour of this 3D VAE can be seen in \autoref{fig:frozen_vae_result}. After applying PCA on the channel dimension (16 channels for CogVideoX), we show that the spatiotemporal encoded latents still have a spatial representation resembling the original video frames. Furthermore, although the 3D VAE was frozen, the encoded sketch latent closely resembles the sketched representation of the colorized video latents. This further shows that, although sketched images exist in a different distribution space from RGB images, the frozen 3D VAE is robust enough to encode tehse sketches, which means that there is no need to fine-tune a specialized sketch encoder, as was done in previous works. 
\subsection{Comparison}
\begin{figure*}
\begin{center}
\includegraphics[width=0.80\linewidth]{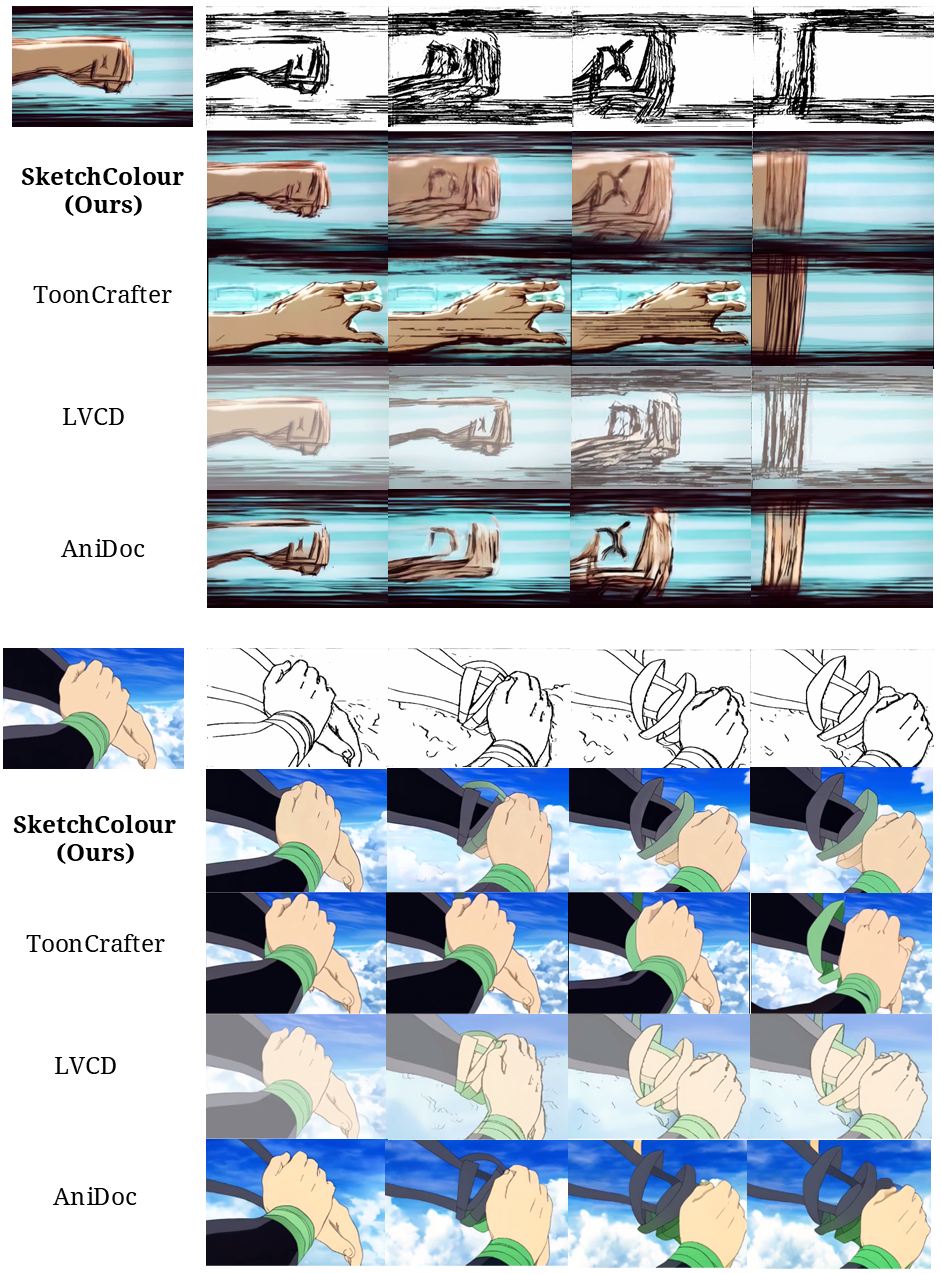}
\end{center}
   \caption{Visual comparison of sketch colorization work with colored first frame as a reference, compared to ToonCrafter \cite{xing2024tooncrafter}, LVCD \cite{huang2024lvcd}, and AniDoc \cite{meng2024anidoc}. }
\label{fig:qualitative_comparison}
\end{figure*}
We evaluate the performance of our model on our test set of 1K randomly sampled clips against three state-of-the-art models for the sketch colorization with frame reference task: LVCD, ToonCrafter, and AniDoc. These three models are all built on diffusion architectures, specifically on a combination of U-Net and ControlNet architectures. With the exception of ToonCrafter, which requires both the colored start and end frames as reference, other models utilize only the colored first frame as reference. 
\paragraph{Quantitative Comparison} Following previous works, we evaluate the quality of the colorized animation in two aspects: video quality and colorization correctness. For video quality, we use Fréchet Video Distance (FVD) \cite{unterthiner2019fvd}, while for colorization correctness we use Mean Squared Color Error (MSCE), which measures mean squared error on color channel, in addition to PSNR, SSIM, LPIPS, each of which measures the similarity of frames using reconstruction metrics. For these metrics, we rescale all of our evaluation videos to the 720 x 480 resolution of the ground truth. For our model, we provide additional result metrics corresponding to videos with frame count matching that of the outputs of LVCD, ToonCrafter, and AniDoc to allow for fair comparison. 

As shown in \autoref{tab:quant_results}, our model has the best result across all metrics, indicating that our model excels in both video quality and colorization correctness. Our videos perform significantly better on PSNR, SSIM and LPIPS, with the score difference against the baselines being half of more of those baselines' standard deviation. We also perform best in terms of MSCE and FVD, with AniDoc trailing behind. It is expected that our model has lower performance with a larger number of frames, as colorizing later frames, which are more different from the colored first frame reference, is harder than colorizing earlier frames. However, our 17 frame model performance remains superior to AniDoc, LVCD, and ToonCrafter with respect to MSCE, PSNR, SSIM, and LPIPS, while on FVD it only loses to AniDoc with a slight margin.

\paragraph{Qualitative Comparison} As shown in \autoref{fig:qualitative_comparison}, our result produces accurate colorization results compared to previous works, adhering closely to the sketch reference while minimizing color bleeding. ToonCrafter's results fail to produce fluid motion and commonly renders deformed objects as shown in \autoref{fig:qualitative_comparison}. LVCD fails to color the video accurately, resulting in a color palette that is substantially duller than that of the reference image, and also suffers from color bleed. Finally, AniDoc captures the overall global colorization result with consistent object modelling, but it still experiences color bleed as \autoref{fig:qualitative_comparison} shows. Our model outperforms the others, showing smooth motion while minimizing object distortion and color bleeding. Additional comparisons and samples can be found on our project page \url{https://bconstantine.github.io/SketchColour/}.
%------------------------------------------------------------------------
\section{Conclusion}
In this paper, we introduce SketchColour, a DiT-based framework for sketch-conditioned 2D animation colorization. By leveraging channel concatenation adapters and LoRA fine-tuning, our approach integrates control signals directly into the diffusion backbone, eliminating the need for a separate ControlNet. Compared to ControlNet, our approach reduces GPU memory requirements and mitigates the latent gap problem. Evaluation on the SAKUGA dataset demonstrates that SketchColour not only surpasses existing U-Net-based video colorization pipelines both with respect to fidelity and temporal consistency, but it also does so with significantly fewer trainable parameters and less training data. Qualitative comparison further highlights our method’s ability to show smooth motion while minimizing object distortion and color bleeding.

%------------------------------------------------------------------------
{\small
\bibliographystyle{ieee_fullname}
\bibliography{egbib}
}

\end{document}